# *Automated Generation of Massive Reasonable Empirical Theorems by Forward Reasoning Based on Strong Relevant Logics - A Solution to the Problem of LLM Pre-training Data Exhaustion*

**Jingde Cheng**

(Professor Emeritus)

Department of Information and Computer Sciences, Saitama University

Saitama, Japan

cheng@aise.ics.saitama-u.ac.jp, jingde.cheng@gmail.com

**Abstract** — Recently, it is often said that the data used for the pre-training of large language models (LLMs) have been exhausted. This paper proposes a solution to the problem: Automated generation of massive reasonable empirical theorems by forward reasoning based on strong relevant logics. In fact, this can be regarded as a part of our approach to the problems of **ATF** (Automated Theorem Finding) and **AKA** (Automated Knowledge Appreciation).

## I. Introduction

Recently, it is often said that the data used for the pre-training of Large Language Models (LLMs) have been exhausted. Thus, we have a problem: How to obtain massive training data of LLMs effectively and efficiently? [11].

The present author considers that a solution to the above problem must satisfy the following basic requirements:

First, the requirement of effectiveness, because the quality of data used for the pre-training of LLMs determines the effectiveness of the pre-training, the solution to the problem must be effective: The obtained data must be correct and not confabulation (i.e., do not include so-called "hallucination").

Second, the requirement of efficiency, because the amount of data used for the pre-training of LLMs is extremely huge, the solution to the problem must be very efficient: a huge amount of data can be generated in a short time.

Third, the requirement of massiveness, because the pre-training of LLMs needs extremely huge data, the solution to the problem must be practicable in the sense of massiveness: Users can generate as much as they want.

This paper proposes a solution that satisfying the above three basic requirements to the problem: Automated generation of massive reasonable empirical theorems by forward reasoning based on strong relevant logics [7,8,20]. In fact, this can be regarded as a part of our approach to the problems of **ATF** (Automated Theorem Finding) [3-7,12,15-19] and **AKA** (Automated Knowledge Appreciation) [10].

The problem of Automated Theorem Finding (**ATF** for short) is one of the 33 basic research problems in automated reasoning which was originally proposed by Wos in 1988 [23,24]. The problem of **ATF** is "What properties can be identified to permit an automated reasoning program to find new and interesting theorems, as opposed to proving conjectured theorems?" [23,24]. The most important and difficult requirement of the problem is that, in contrast to proving conjectured theorems supplied by the user, it asks for the criteria that an automated reasoning program can use to find some theorems in a field that must be evaluated by theorists of the field as new and interesting theorems [3-7,12,15-19]. The significance of solving the problem is obvious because an automated reasoning program satisfying the requirement can provide great assistance for scientists in various fields [3-7,12,15-19].

The present author proposed in 2015 a novel research direction: Automated Knowledge Appreciation (**AKA** for short), that intents to establish a general systematic methodology and develop automated tools to expand our knowledge and increase its value automatically [10]. By "Knowledge Appreciation", we mean the following two points: first, to expand our knowledge based on our known knowledge, and second, compared to the value of our known knowledge, the value of our expanded knowledge increased. Note that the above definition of Knowledge Appreciation implicitly require there should be some evaluation criterion about the value of knowledge, even if the evaluation criterion may be dependent on the cognitive subject.

To find new theorems and to appreciate new knowledge, we have to generate hopeful reasonable candidates for the new theorems and new knowledge as much as possible, thus, this leads the present author to propose our solution to the problem of LLM pre-training data exhaustion.

The rest of the paper is organized as follows: Section 2 presents basic notions and notations used in this paper, Section 3 discusses logical basis for **ATF** and **AKA**, Section 4 present our relevant reasoning approach to the problem of LLM pre-training data exhaustion, and concluding remarks are given in Section 5.

## II. BASIC NOTIONS AND NOTATIONS

*Reasoning* is the process of drawing new conclusions from given premises, which are already known facts or previously assumed hypotheses to provide some evidence for the conclusions (Note that how to define the notion of 'new' formally and satisfactorily is still a difficult open problem until now). Therefore, reasoning is intrinsically ampliative, i.e., it has the function of enlarging or extending some things, or adding some things to what is already known or assumed. In general, a reasoning consists of a number of arguments in some order.

An *argument* is a set of statements (or declarative sentences) of which one statement is intended as the conclusion, and one or more statements, called "premises," are intended to provide some evidence for the conclusion. An argument is a conclusion standing in relation to its supporting evidence. In an argument, a claim is being made that there is some sort of evidential relation between its premises and its conclusion: the conclusion is supposed to follow from the premises, or equivalently, the premises are supposed to entail the conclusion. Therefore, the correctness of an argument is a matter of the connection between its premises and its conclusion, and concerns the strength of the relation between them (Note that the correctness of an argument depends neither on whether the premises are really true or not, nor on whether the conclusion is really true or not). Thus, there are some fundamental questions: What is the criterion by which one can decide whether the conclusion of an argument or a reasoning really does follow from its premises or not? Is there the only one criterion, or are there many criteria? If there are many criteria, what are the intrinsic differences between them? It is logic that deals with the validity of argument and reasoning in a general theory.

A *logically valid reasoning* is a reasoning such that its arguments are justified based on some logical validity criterion provided by a logic system in order to obtain correct conclusions (Note that here the term 'correct' does not necessarily mean 'true'). Today, there are so many different logic systems motivated by various philosophical considerations. As a result, a reasoning may be valid on one logical validity criterion but invalid on another. For example, the classical account of validity, which is one of fundamental principles and assumptions underlying classical mathematical logic and its various conservative extensions, is defined in terms of truth-preservation (in some certain sense of truth) as: an argument is valid if and only if it is impossible for all its premises to be true while its conclusion is false. Therefore, a classically valid reasoning must be truth-preserving. On the other hand, for any correct argument in scientific reasoning as well as our everyday reasoning, its premises must somehow be relevant to its conclusion, and vice versa. The relevant account of validity is defined in terms of relevance as: for an argument to be valid there must be some connection of meaning, i.e., some relevance, between its premises and its conclusion. Obviously, the relevance between the premises and conclusion of an argument is not accounted for by the classical logical validity criterion, and therefore, a classically valid reasoning is not necessarily relevantly valid.

***Proving*** is the process of finding a justification for an explicitly specified statement from given premises, which are already known facts or previously assumed hypotheses to provide some evidence for the specified statement. A proof is a description of a found justification. A ***logically valid proving*** is a proving such that it is justified based on some logical validity criterion provided by a logic system in order to obtain a correct proof.

The most intrinsic difference between reasoning and proving is that the former is intrinsically prescriptive and predictive while the latter is intrinsically descriptive and non-predictive. The purpose of reasoning is to find some new conclusion previously unknown or unrecognized, while the purpose of proving is to find a justification for some specified statement previously given. Proving has an explicitly given target as its goal while reasoning does not.

***Logic*** deals with what entails what or what follows from what, and aims at determining which are the correct conclusions of a given set of premises, i.e., to determine which arguments are correct and/or valid. Therefore, the most essential and central concept in logic is the ***logical consequence relation*** that relates a given set of premises to those conclusions, which validly follow from the premises. To define a logical consequence relation is nothing else but to provide a logical validity criterion by which one can decide whether the conclusion of an argument or a reasoning really does follow from its premises or not. Moreover, to answer the question what is the correct conclusion of given premises, we have to answer the question: correct for what? Based on different philosophical motivations, one can define various logical consequence relations and therefore establish various logic systems.

In logic, a sentence in the form of 'if ... then ...' is usually called a ***conditional proposition*** or simply ***conditional*** which states that there exists a relation of sufficient condition between the 'if' part and the 'then' part of the sentence. In general, a conditional must concern two parts which are connected by the connective 'if ... then ...' and called the ***antecedent*** and the ***consequent*** of that conditional, respectively. The truth of a conditional depends not only on the truth of its antecedent and consequent but also, and more essentially, on a necessarily relevant and conditional relation between them. The notion of conditional plays the most essential role in reasoning because any reasoning form must invoke it, and therefore, it is historically always the most important subject studied in logic and is regarded as "the heart of logic" [1].

When we study and use logic, the notion of conditional may appear in both the object logic (i.e., the logic we are studying) and the meta-logic (i.e., the logic we are using to study the object logic). In the object logic, there usually is a connective in its formal language to represent the notion of conditional, and the notion of conditional, usually represented by a meta-linguistic symbol, is also used for representing a logical consequence relation in its proof theory or model theory. On the other hand, in the meta-logic, the notion of conditional, usually in the form of natural language, is used for defining various meta-notions and describing various meta-theorems about the object logic.

From the viewpoint of object logic, there are two classes of conditionals. One class is ***empirical conditionals*** and the other class is ***logical conditionals***. For a logic, a conditional is called an empirical conditional of the logic if its truth-value, in the sense of that logic, depends on the contents of its antecedent and consequent and therefore cannot be determined only by its abstract form (i.e., from the viewpoint of that logic, the relevant relation between the antecedent and the consequent of that conditional is regarded to be empirical); a conditional is called a logical conditional of the logic if its truth-value, in the sense of that logic, depends only on its abstract form but not on the contents of its antecedent and consequent, and therefore, it is considered to be universally true or false (i.e., from the viewpoint of that logic, the relevant relation between the antecedent and the consequent of that conditional is regarded to be logical). A logical conditional that is considered to be universally true, in the sense of that logic, is also called an ***entailment*** of that logic. Indeed, the most intrinsic difference between various different logic systems is to regard what class of conditionals as entailments, as Diaz pointed out: "The problem in modern logic can best be put as follows: can we give an explanation of those conditionals that represent an entailment relation?" [13]

A ***formal logic system* L** consists of a formal language, called the object language and denoted by F(**L**), which is the set of all well-formed formulas of **L**, and a logical consequence relation, denoted by meta-linguistic symbol $|\!-_\mathbf{L}$, such that for $P \subseteq$ F(**L**) and $c \in$ F(**L**), $P |\!-_\mathbf{L} c$ means that within the framework of **L**, *c* is a valid conclusion of premises *P*, or given *P* as premises, *c* as a valid conclusion follows from *P*.

For a formal logic system (F(**L**), $|\!-_\mathbf{L}$), a ***logical theorem*** *t* is a formula of **L** such that $\phi |\!-_\mathbf{L} t$ where $\phi$ is the empty set. We use Th(**L**) to denote the set of all logical theorems of **L**. Th(**L**) is completely determined by the logical consequence relation $|\!-_\mathbf{L}$. According to the representation of the logical consequence relation of a logic system, the logic can be represented as a Hilbert style axiomatic system, a Gentzen natural deduction system, a Gentzen sequent calculus system, or other type of formal system.

A formal logic system **L** is said to be ***explosive*** if and only if $\{A, \neg A\} |\!-_\mathbf{L} B$ for any two different formulas *A* and *B*; **L** is said to be ***paraconsistent*** if and only if it is not explosive.

Let (F(**L**), $|\!-_\mathbf{L}$) be a formal logic system and $P \subseteq$ F(**L**) be a non-empty set of sentences (i.e. closed well-formed formulas). A ***formal theory*** with premises *P* based on **L**, called an ***L-theory with premises P*** and denoted by $T_\mathbf{L}(P)$, is defined as $T_\mathbf{L}(P) =_{df}$ Th(**L**) $\cup$ $Th_\mathbf{L}^e(P)$, and $Th_\mathbf{L}^e(P) =_{df} \{et \mid P |\!-_\mathbf{L} et$ and $et \notin$ Th(**L**)$\}$ where Th(**L**) and $Th_\mathbf{L}^e(P)$ are called the ***logical part*** and the ***empirical part*** of the formal theory, respectively, and any element of $Th_\mathbf{L}^e(P)$ is called an ***empirical theorem*** of the formal theory.

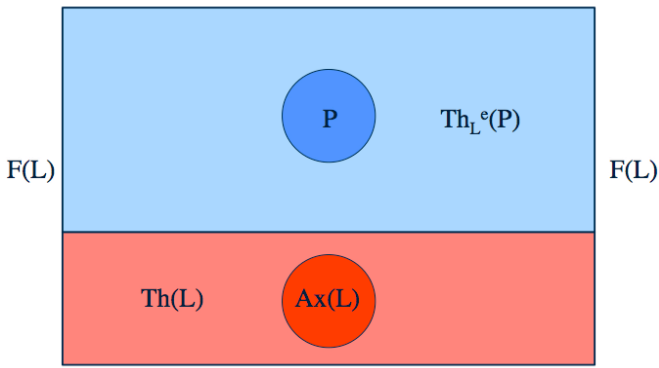

Figure 1   Formal theory

Figure 1 shows a formal theory. We can see that a formal theory with premises *P* based on logic **L** in fact classify all logical formulas of **L** into three parts: the logical part (red part), the empirical part (blue part), and the non-theorem part (white part).

A formal theory $T_\mathbf{L}(P)$ is said to be ***directly inconsistent*** if and only if there exists a formula *A* of **L** such that both $A \in P$ and $\neg A \in P$ hold. A formal theory $T_\mathbf{L}(P)$ is said to be ***indirectly inconsistent*** if and only if it is not directly inconsistent but there exists a formula *A* of **L** such that both $A \in T_\mathbf{L}(P)$ and $\neg A \in T_\mathbf{L}(P)$. A formal theory $T_\mathbf{L}(P)$ is said to be ***consistent*** if and only if it is neither directly inconsistent nor indirectly inconsistent. A formal theory $T_\mathbf{L}(P)$ is said to be ***explosive*** if and only if $A \in T_\mathbf{L}(P)$ for any $A \in$ F(**L**); $T_\mathbf{L}(P)$ is said to be ***paraconsistent*** if and only if it is not explosive. An explosive formal theory is not useful at all. Therefore, any meaningful formal theory should be paraconsistent. Note that if a formal logic system **L** is explosive, then any directly or indirectly inconsistent **L**-theory $T_\mathbf{L}(P)$ must be explosive.

In a formal logic system with a primitive connective to represent the notion of conditional, a formula may have a structure of deeply nested conditional. With the depth of nested conditional, a formal logic system can be divided into different fragments about different degree of nested conditional.

For a formal logic system where the notion of conditional is represented by a primitive connective, say '⇒', a formula is called a ***zero degree formula*** if and only if there is no occurrence of ⇒ in it; a formula of the form *A*⇒*B* or *Q*(*A*⇒*B*) where *Q* is the quantifier prefix of *A*⇒*B* is called a ***first degree conditional*** if and only if both *A* and *B* are zero degree formulas; a formula *A* is called a ***first degree formula*** if and only if it satisfies the one of the following conditions: (1) *A* is a first degree conditional, (2) *A* is in the form +*B* or *Q*(+*B*) where + is a one-place connective such as negation and so on and *Q* is the quantifier prefix of +*B* such that *B* is a first degree formula, and (3) *A* is in the form *B\*C* or *Q*(*B\*C*) where \* is a non-conditional two-place connective such as conjunction or disjunction and so on and *Q* is the quantifier prefix of *B\*C* such that both of *B* and *C* are first degree formulas, or one of *B* and *C* is a first degree formula and another is a zero degree formula.

The notion of degree of nested conditional can also be generally defined as follows. Let $A, B, C$ are formulas. The degree of conditional in $A$, denoted by $D_\Rightarrow(A)$, is inductively defined as follows: (1) $D_\Rightarrow(A) = 0$ if and only if there is no occurrence of $\Rightarrow$ in $A$, (2) if $A$ is a conditional of the form $B \Rightarrow C$, then $D_\Rightarrow(A) = \max\{D_\Rightarrow(B), D_\Rightarrow(C)\} + 1$, (3) if $A$ is in the form $+B$ where $+$ is a one-place connective such as negation and so on, then $D_\Rightarrow(A) = D_\Rightarrow(B)$, (4) if $A$ is in the form $B*C$ where $*$ is a non-conditional two-place connective such as conjunction or disjunction and so on, then $D_\Rightarrow(A) = \max\{D_\Rightarrow(B), D_\Rightarrow(C)\}$, and (5) if $A$ is in the form $QB$ where $Q$ is the quantifier prefix of $B$, then $D_\Rightarrow(A) = D_\Rightarrow(B)$. Let $k$ be a natural number. A formula $A$ is called a $k^{th}$ *degree formula* if and only if $D_\Rightarrow(A) = k$, in particular, $A$ is called a $k^{th}$ *degree conditional* if it is a conditional.

Let $(F(L), \vdash_L)$ be a formal logic system and $k$ be a natural number. The $k^{th}$ *degree fragment* of $L$, denoted by $Th^k(L)$, is a set of logical theorems of $L$ which is inductively defined as follows (in the terms of Hilbert style axiomatic system): (1) if $A$ is an axiom of $L$ and $D_\Rightarrow(A) \leq k$, then $A \in Th^k(L)$, (2) if $A$ is the result of applying an inference rule of $L$ to some members of $Th^k(L)$ and $D_\Rightarrow(A) \leq k$, then $A \in Th^k(L)$, and (3) Nothing else are members of $Th^k(L)$, i.e., only those obtained from repeated applications of (1) and (2) are members of $Th^k(L)$. Note that the $k^{th}$ degree fragment of logic $L$ does not necessarily include all $k^{th}$ degree logical theorems of $L$ because it is possible for $L$ that deductions of some $k^{th}$ degree logical theorems of $L$ must invoke those logical theorems whose degrees are higher than $k$. On the other hand, the following holds obviously: $Th^0(L) \subset Th^1(L) \subset ... Th^{k-1}(L) \subset Th^k(L) \subset Th^{k+1}(L) \subset ...$.

Let $(F(L), \vdash_L)$ be a formal logic system, $P \subset F(L)$ ($P \neq \phi$), and $k$ and $j$ be two natural numbers. A formula $A$ is said to be $j^{th}$-*degree-deducible from $P$ based on $Th^k(L)$* if and only if there is a finite sequence of formulas $f_1, ..., f_n$ such that $f_n = A$ and for all $i$ ($i \leq n$), (1) $f_i \in Th^k(L)$, or (2) $f_i \in P$, or (3) $f_i$ is the result of applying an inference rule to some members $f_{j1}, ..., f_{jm}$ ($j1, ..., jm < i$) of the sequence and $D_\Rightarrow(f_i) \leq j$. The set of all formulas which are $j^{th}$-degree-deducible from $P$ based on $Th^k(L)$ is called the $j^{th}$ *degree fragment of the formal theory with premises $P$ based on $Th^k(L)$*, denoted by $T^j_{Th^k(L)}(P)$. A formula is said to be $j^{th}$-*degree-deductive from $P$ based on $Th^k(L)$* if and only if it is $j^{th}$-degree-deducible from $P$ based on $Th^k(L)$ but not $(j-1)^{th}$-degree-deducible from $P$ based on $Th^k(L)$. Note that in the above definitions, we do not require $j \leq k$. The notion of $j^{th}$-degree-deductive can be regarded as a metric to measure the difficulty of deducing an empirical theorem from given premises $P$ based on logic $L$. The difficulty is relative to the complexity of problem being investigated as well as the strength of logic $L$.

The notion of degree of nested conditional can be generally extended to other logic connectives and modal operators as follows: Let $\theta$ be an arbitrary $n$-ary ($1 \leq n$) connective or modal operator of logic $L$ and $A$ be a formula of $L$, the degree of $\theta$ in $A$, denoted by $D_\theta(A)$, is defined as follows: (1) $D_\theta(A) = 0$ if and only if there is no occurrence of $\theta$ in $A$, (2) if $A$ is in the form $\theta(a_1, ..., a_n)$ where $a_1, ..., a_n$ are formulas, then $D_\theta(A) = \max\{D_\theta(a_1), ..., D_\theta(a_n)\} + 1$, (3) if $A$ is in the form $\sigma(a_1, ..., a_n)$ where $\sigma$ is a connective or modal operator different from $\theta$ and $a_1, ..., a_n$ are formulas, then $D_\theta(A) = \max\{D_\theta(a_1), ..., D_\theta(a_n)\}$, and (4) if $A$ is in the form $QB$ where $B$ is a formula and $Q$ is the quantifier prefix of $B$, then $D_\theta(A) = D_\theta(B)$.

The notion of degree of logic fragment about conditional can also be generally extended to other logic connectives and/or modal operators as follows: Let $\theta_1, ..., \theta_n$ be logic connectives and/or modal operators of logic $L$ and $k_1, ..., k_n$ be natural numbers, the fragment of $L$ about $\theta_1, ..., \theta_n$ and their degrees $k_1, ..., k_n$, denoted by $Th^{(\theta_1, k_1, ..., \theta_n, k_n)}(L)$, is a set of logical theorems of $L$ which is inductively defined as follows (in the terms of Hilbert style axiomatic system): (1) if $A$ is an axiom of $L$ and $D_{\theta_1}(A) \leq k_1, ..., D_{\theta_n}(A) \leq k_n$, then $A \in Th^{(\theta_1, k_1, ..., \theta_n, k_n)}(L)$, (2) if $A$ is the result of applying an inference rule of $L$ to some members of $Th^{(\theta_1, k_1, ..., \theta_n, k_n)}(L)$ and $D_{\theta_1}(A) \leq k_1, ..., D_{\theta_n}(A) \leq k_n$, then $A \in Th^{(\theta_1, k_1, ..., \theta_n, k_n)}(L)$, and (3) Nothing else are members of $Th^{(\theta_1, k_1, ..., \theta_n, k_n)}(L)$.

The notion of the $j^{th}$ degree fragment (about conditional) of the formal theory with premises $P$ based on $Th^k(L)$ can also be generally extended to the case of $Th^{(\theta_1, k_1, ..., \theta_n, k_n)}(L)$ by specifying degrees $j_1, …, j_n$ for logic connectives and/or modal operators $θ_1, …, θ_n$ respectively.

### III. LOGICAL BASIS FOR ATF AND AKA

Using a formal theory based on some logic **L** as a formal representation, if we regard our known knowledge as premises *P*, then automated knowledge expansion in our problem of **AKA** can be regarded as how to find new empirical theorems in a formal theory (**L**-theory) with premises *P* automatically. In other words, based on formal theory representation, we can formulize the problem of automated knowledge expansion as the problem of **ATF**.

Now, the next problem is which logic system **L** can underlie **ATF** and **AKA** satisfactorily? The present author considers that the fundamental logic system underlying **ATF** and **AKA** must satisfy the following three essential requirements.

First, as a general logical criterion for the validity of reasoning as well as proving, the fundamental logic must be able to underlie relevant reasoning as well as truth-preserving reasoning in the sense of conditional, i.e., for any reasoning based on the logic to be valid, if its premises are true in the sense of conditional, then its conclusion must be relevant to the premises and must be true in the sense of conditional. Logic is the study of the methods and principles used to distinguish correct reasoning from incorrect reasoning, and the notion of a conditional is the heart of logic. Because there is no reasoning that does not invoke the notion of conditional, this requirement is primarily important.

Second, the fundamental logic must be able to underlie ampliative reasoning, i.e., for any reasoning based on the logic to be valid, the truth of conclusion of the reasoning should be recognized after the completion of the reasoning process but not be invoked in deciding the truth of premises of the reasoning. From the viewpoint of regarding reasoning as the process of drawing new conclusions from given premises, any meaningful reasoning must be ampliative but not circular and/or tautological.

Third, the fundamental logic must be able to underlie paracomplete reasoning and paraconsistent reasoning, i.e., the conclusion of the reasoning may not be the negation of a sentence even if the sentence is not a conclusion of the premises of that reasoning, and also may not be an arbitrary sentence even if the premises of that reasoning is inconsistent. In particular, the so-called principle of Explosion that everything follows from a contradiction should not be accepted by the logic as a valid principle. In general, our knowledge about a domain as well as a scientific discipline may be incomplete and/or inconsistent in many ways, i.e., it gives us no evidence for deciding the truth of either a proposition or its negation, and/or it directly or indirectly includes some contradictions. Therefore, reasoning with incomplete and/or inconsistent knowledge is the rule rather than the exception in our everyday lives and almost all scientific disciplines.

For any valid reasoning with correct premises based on a fundamental logic system, we can directly accept its conclusion as correct one without evaluation, ONLY IF the fundamental logic system satisfies the above three essential requirements.

Classical mathematical logic (CML for short) was established in order to provide formal languages for describing the structures with which mathematicians work, and the methods of proof available to them; its principal aim is a precise and adequate understanding of the notion of mathematical proof. CML was established based on a number of fundamental assumptions. Among them, the most characteristic one is the classical account of validity (CAV for short) that is the logical validity criterion of CML by which one can decide whether the conclusion of an argument or a reasoning really does follow from its premises or not in the framework of CML. However, since the relevance between the premises and conclusion of an argument is not accounted for by the classical validity criterion, a reasoning based on CML is not necessarily relevant. On the other hand, in CML the notion of conditional, which is intrinsically intensional but not truth-functional, is represented by the notion of material implication, which is intrinsically an extensional truth-function. This leads to the problem of 'implicational paradoxes' [1, 2, 13, 14].

CML cannot satisfy any of the above three essential requirements for the fundamental logic system because of the following facts: a reasoning based on CML is not necessarily relevant; the classical truth-preserving property of a reasoning based on CML is meaningless in the sense of conditional; a reasoning based on CML must be circular and/or tautological but not ampliative; reasoning under inconsistency is impossible within the framework of CML. These facts are also true to those classical conservative extensions or non-classical alternatives of CML where the classical account of validity is adopted as the logical validity criterion and the notion of conditional is directly or indirectly represented by the material implication.

Traditional relevant (or relevance) logics were constructed during the 1950s in order to find a mathematically satisfactory way of grasping the elusive notion of relevance of antecedent to consequent in conditionals, and to obtain a notion of implication which is free from the so-called 'paradoxes' of material and strict implication [1, 2, 13, 14]. Some major traditional relevant logic systems are 'system E of entailment', 'system R of relevant implication', and 'system T of ticket entailment'. A major characteristic of the relevant logics is that they have a primitive intensional connective to represent the notion of conditional (entailment) and their logical theorems include no implicational paradoxes. The underlying principle of the relevant logics is the relevance principle, i.e., for any entailment provable in E, R, or T, its antecedent and consequent must share a propositional variable. Variable-sharing is a formal notion designed to reflect the idea that there be a meaning-connection between the antecedent and consequent of an entailment. It is this relevance principle that excludes those implicational paradoxes from logical axioms or theorems of relevant logics. Also, since the notion of entailment is represented in the relevant logics by a primitive intensional connective but not an extensional truth-function, a reasoning based on the relevant logics is ampliative but not circular and/or tautological. Moreover, because the relevant logics reject the principle of Explosion, they can certainly underlie paraconsistent reasoning. However, because the relevance principle cannot exclude those conditionals whose antecedent including unnecessary and needless conjuncts or whose consequent including unnecessary and needless disjuncts, logical theorems of traditional relevant (or relevance) logics still include some conjunction-implicational paradoxes and disjunction-implicational paradoxes [7,8].

In order to establish a satisfactory logic calculus of conditional to underlie relevant reasoning, the present author has proposed some strong relevant (or relevance) logics (SRLs for short), named Rc, Ec, and Tc [7,8,20]. The SRLs require that the premises of an argument represented by a conditional include no unnecessary and needless conjuncts and the conclusion of that argument includes no unnecessary and needless disjuncts. As a modification of traditional relevant logics R, E, and T, SRLs Rc, Ec, and Tc rejects all conjunction-implicational paradoxes and disjunction-implicational paradoxes in R, E, and T, respectively. What underlies the SRLs is the strong relevance principle: for any theorem of Rc, Ec, or Tc, every propositional variable in the theorem occurs at least once as an antecedent part and at least once as a consequent part. Since the SRLs are free of not only implicational paradoxes but also conjunction-implicational and disjunction-implicational paradoxes, in the framework of strong relevant logics, if a reasoning is valid, then both the relevance between its premises and its conclusion and the validity of its conclusion in the sense of conditional can be guaranteed in a certain sense of strong relevance [7,8,20].

SRLs can satisfy the above three essential requirements for the fundamental logic system to underlie **ATF** and **AKA**. Probably, SRLs is the only family of logics that can satisfy the above three essential requirements at present.

On the other hand, the above three essential requirements for the fundamental logic system are necessary but not complete, if we deal with our knowledge and its update about a changing world [9]. This naturally requires that the fundamental logic system must be able to underlie temporal, three-dimensional spatial, and three-dimensional spatio-temporal reasoning. The present author has proposed some temporal, three-dimensional spatial, and three-dimensional spatio-temporal conservative extensions of SRLs as shown in Figure 2 [8].

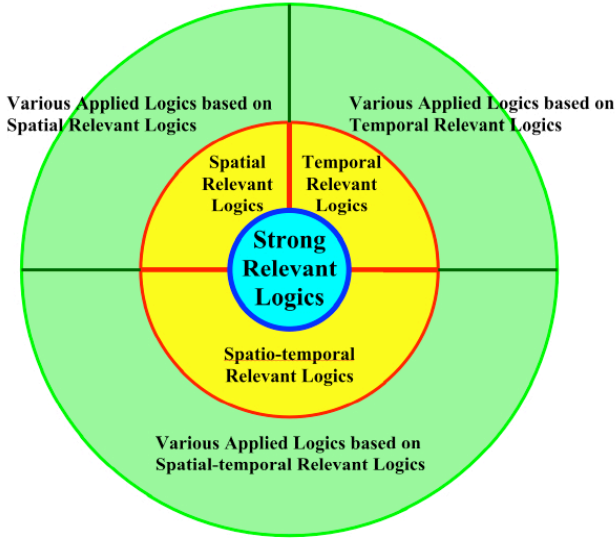

Figure 2   SRLs and its Conservative Extensions

## IV. A Relevant Reasoning Approach to the problem of LLM pre-training data exhaustion

Based on our experiences of development and applications of EnCal [5], we have developed a forward reasoning engine with general-purpose, named "FreeEnCal", which can interpret and perform inference rules defined and given by its users, draw fragments of various logic systems specified by its users, draw empirical theorems of various formal theories constructed based on various logic systems, and perform deductive, inductive, and abductive reasoning automatically [12].

FreeEnCal can be used as a ready-made forward reasoning engine serving as a core and fundamental component in various advanced knowledge-based systems as well as an alone forward reasoning engine with general-purpose. In particular, FreeEnCal can be used as a core and fundamental component in various hybrid intelligent systems with multiparadigms because it can provide its users with diverse, flexible, and powerful ways to define and perform various reasoning tasks. FreeEnCal can be used as automated forward reasoning engine in **ATF** and **AKA**.

For **ATF** in formal theories based on SRLs, we have proposed a systematic methodology and shown the effectiveness of our methodology by some case studies [3-7,12,18,19,21].

As shown in Figure 3, finding empirical theorems in the formal theory of a target field by FreeEnCal Based on SRLs has two stages. First, we select a SRL logic fragment, as input give its axioms, inference rules, and degree numbers of connectives to FreeEnCal, and let FreeEnCal draw out that fragment. Second, we use the SRL logic fragment as logic basis, as input give the fragment, inference rules, degree numbers of connectives, empirical axioms of the target field to FreeEnCal, and let FreeEnCal draw out empirical theorems of that target field.

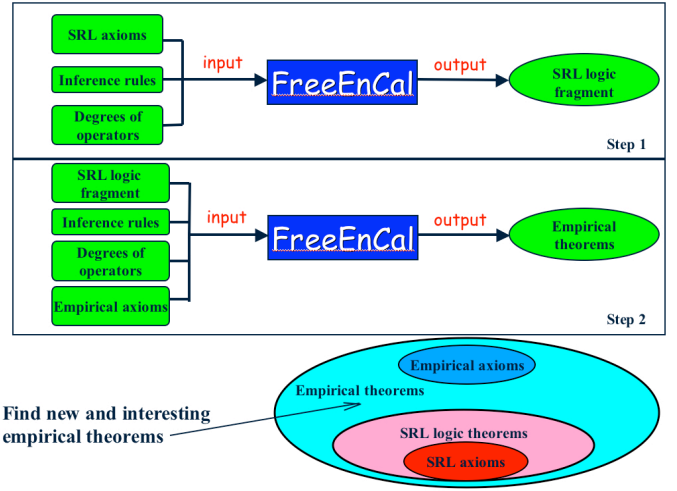

Figure 3   **ATF** in formal theories by FreeEnCal based on SRL

We can perform automated knowledge expansion in a ready-for-customization way, i.e., we prepare necessary SRL logic fragments and various basic mathematical formal theories (see Figure 4 for an example of SRL-theory of axiomatic set theory, Figure 5 for an example of SRL-theory of number theory, and Figure 6 for other examples) as more as possible, and make ready for customization / order from users. Thus, any user want to expand his/her knowledge just need to submit a set of premises (i.e, empirical axioms, facts, assumptions / hypotheses), select a suitable formal theory, and let FreeEnCal draw out empirical theorems of the target formal theory.

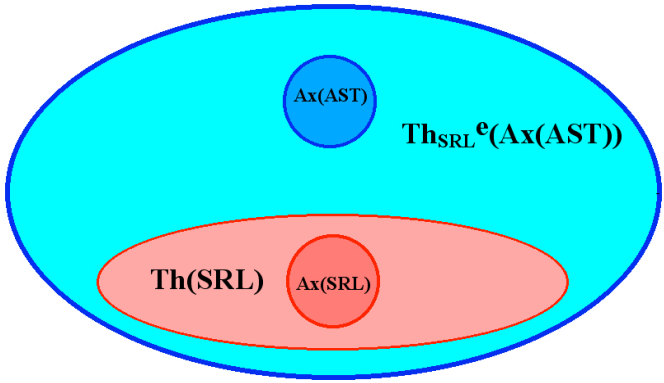

Figure 4   SRL-theory of axiomatic set theory

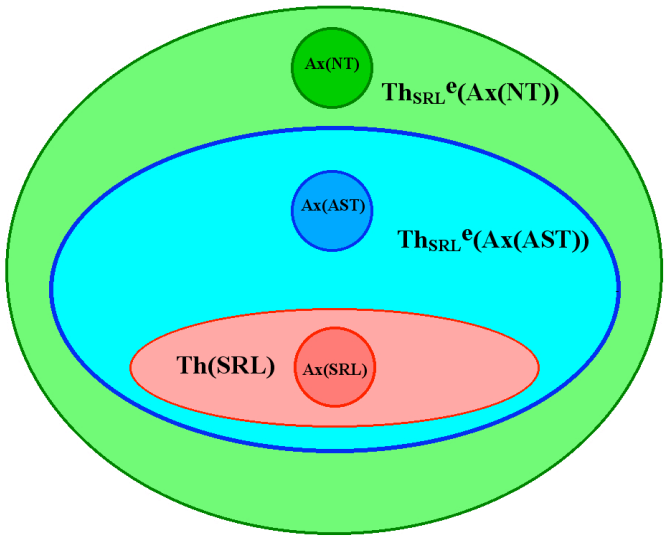

Figure 5   SRL-theory of number theory

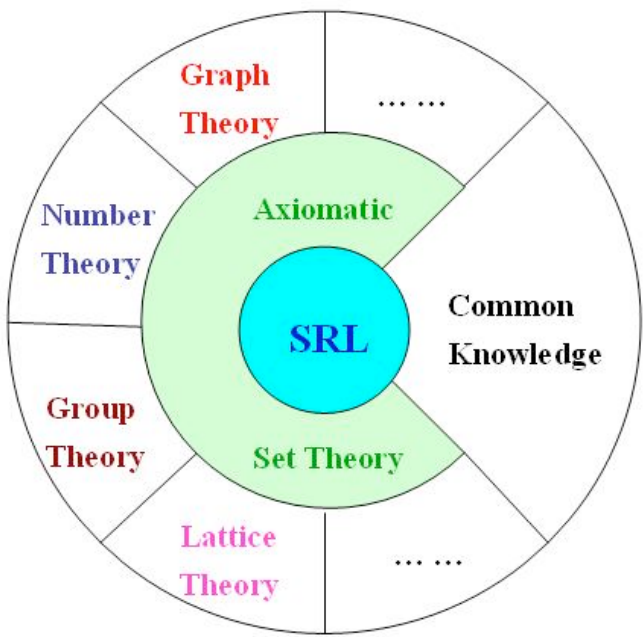

Figure 6   Various SRL-Theories

For those users whose knowledge are based on other basic theories such as physics, chemistry, computer science, artificial intelligence, software engineering, knowledge engineering, information security engineering, and so on, it is necessary to prepare those basic formal theories at first.

Of course, maybe some users do not need any basic mathematical formal theory and other formal theories as the basis of their knowledge. In such case, the users just need to submit their known knowledge as premise, select a suitable SRL logic fragment, let FreeEnCal draw out empirical theorems of the target formal theory, and expect to find some new and interesting knowledge in the target formal theory.

The steps to automatically generate massive reasonable empirical theorems by forward reasoning based on SRLs are as follows:

1. According to the target domain, select a suitable SRL system as the fundamental logic system.

2. Determine the allowable nesting degree of the logical connectives for generating the fundamental logic system fragments according to the required data scale requirements.

3. Use an automatic forward reasoning engine (such as FreeEnCal developed by the present author's lab) to generate the fundamental logic system fragments.

4. Determine the allowable nesting degree of the logical connectives for generating empirical theorems according to the required data scale requirements.

5. Prepare premises of empirical facts and empirical axioms.

6. Starting from the given premises, based on the fundamental logic system fragments obtained in the third step, use an automatic forward reasoning engine (such as FreeEnCal developed by the present author's lab) to generate a large number of empirical theorems.

For example, as shown in Figure 7, an expert of RSA public-key cryptosystem, which is based on number theory, may submit his known knowledge about RSA public-key cryptosystem as premise, select a suitable SRL logic fragment, let FreeEnCal draw out empirical theorems of the target formal theory, SRL-theory of RSA public-key cryptosystem $Th_{SRL}^e(RSA)$.

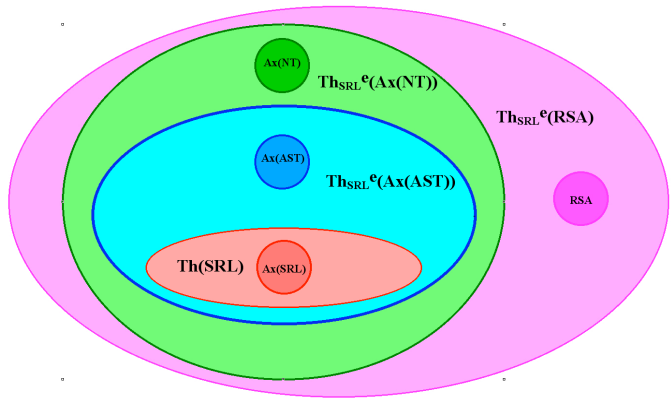

Figure 7   SRL-Theory of RSA public-key cryptosystem

Because the fifth step above needs to transform the data usually expressed in natural language into logical formula expressions, and the generated empirical theorems are also logical formula expressions, the pre-training corpus data used as the large language model also needs to be transformed into natural language expressions first, so a natural language <=> logical formula bidirectional automatic transformation tool is also needed [22].

## V. CONCLUDING REMARKS

We have proposed a solution to the problem of LLM pre-training data exhaustion: Automated generation of massive reasonable empirical theorems by forward reasoning based on strong relevant logics.

As long as the original corpus data as the premise of empirical facts and empirical axioms are pure, then the new empirical theorem obtained in the above way must also be pure as corpus data; if the original corpus data is not pure but is polluted to some extent, then the degree of pollution of the new corpus data obtained in the above way will not exceed that of the original corpus data, that is, the above method is "clean", which will neither clarify the pollution in the original corpus data nor increase new pollution in the newly generated corpus data. The key to this "cleanliness" lies in the choice of the correct fundamental logic system, which will not produce "garbage" when performing forward reasoning. The correct fundamental logic system, at present, should be one of strong relevant logics.